\documentclass{article}

\usepackage[preprint]{neurips_2022}
\usepackage{amsmath}
\usepackage[dvipsnames]{xcolor}
\usepackage{amsfonts}
\usepackage{graphicx}
\usepackage{caption} 
\usepackage[utf8]{inputenc} 
\usepackage[T1]{fontenc}    
\usepackage{hyperref}       
\usepackage{url}            
\usepackage{booktabs}       
\usepackage{amsfonts}       
\usepackage{nicefrac}       
\usepackage{microtype}      
\usepackage{xcolor}         

\title{Learn to explain yourself, when you can:\\ Equipping Concept Bottleneck Models with the ability to abstain on their concept predictions}

\author{%
  Joshua Lockhart \\
  J.P. Morgan AI Research\\
\texttt{joshualockhart@gmail.com} \\
  \And
   Daniele Magazzeni \\
   J.P. Morgan AI Research\\
  \And
  Manuela Veloso \\
   J.P. Morgan AI Research\\
}

\begin{document}

\maketitle
\newcommand{\todo}[1]{\textcolor{red}{#1}}

\begin{abstract}
The Concept Bottleneck Models (CBMs) of \cite{CBMs} provide a means to ensure that a neural network based classifier bases its predictions solely on human understandable concepts. The concept labels, or \textit{rationales} as we refer to them, are learned by the concept labeling component of the CBM. Another component learns to predict the target classification label from these predicted concept labels. Unfortunately, these models are heavily reliant on human provided concept labels for each datapoint. To enable CBMs to behave robustly when these labels are not readily available, we show how to equip them with the ability to abstain from predicting concepts when the concept labeling component is uncertain. In other words, our model learns to provide rationales for its predictions, but only whenever it is sure the rationale is correct.
\end{abstract}

\section{Introduction} 
Neural networks can achieve tremendously accurate results on a variety of classification tasks (see, among others, \cite{schmidhuber2015deep}, \cite{goodfellow2016deep},\cite{vaswani2017attention}, \cite{radford2021learning}, \cite{dosovitskiy2020image}). However, neural networks are often referred to as \textit{black box} models since it is difficult to understand \textit{why} they produced a particular prediction from examining their internal workings. Machine learning researchers must be mindful of the fact that their models may (eventually) form part of a production software system. These systems will have users, and the predictions made by the model will impact human beings in some capacity. Thus, the \textit{accuracy} of the model should be recognised for what it is as (i) only one dimension of a broader set of concerns and desiderata for that model, based on the context in which the model predictions will be used; (ii) often only a limited view on what it means for a model to be ``correct'' about a prediction. It is not difficult to come up with requirements beyond accuracy for a model's predictions; one that we focus on in this work is that the model's prediction should be justifiable, or to use the terminology emerging in recent discourse \textit{interpretable}, or \textit{explainable}. Point (ii) above is also important to consider, since intuition around the accuracy of a deployed model's predictions can be difficult to attain: after performing well on a test set of datapoints, a model is normally put into some trial rollout phase where the model developers determine if it is performing well according to some appropriate metrics. It is difficult to judge if the accuracy metrics defined by the model developers capture the salient aspects of the software system as a whole.

Much of the latter discussion lies out of scope of this work, but we concern ourselves with a small part of the overall field of explainable AI: models that are capable of providing justifications for their predictions.
Explainable AI, as surveyed by \cite{dovsilovic2018explainable}, and \cite{arrieta2020explainable}, has expanded rapidly in the past decade as practitioners and end users have started demanding more of machine learning models.
As touched on above, \textit{accuracy} is only part of the story: when a model is deemed accurate, what is often meant is that the model performs well on a set of held out datapoints (test set), and performs well (so far!) in terms of the \textit{model owner's judgement} of what the correct labels are. One of the criteria explainable AI demands is that a model's decisions should be as transparent as possible. If the inner workings of a model are not transparent, then ethically, the model owner should at least be able to stand by some approximation of a rationale for the decision, and be confident that the model is basing its decision on something reasonable. Indeed, \cite{rudin2019stop} disagrees with using black box models at all for high stakes decision making .

In this work we consider \textit{concept based classification}, a burgeoning field of research on classifiers that are trained to base their decisions on human-provided concept labels. Key to this area are the Concept Bottleneck Models (CBMs) of \cite{CBMs}, a class of neural network based models that first project their input data into a \textit{concept activation vector} (CAV), upon which all subsequent classifications are based. In this way, the predicted CAV is a kind of explanation, or justification, of the predicted target label. Additionally, the bottleneck aspect of these models, that all decision making for the target label predictions is in terms of the CAV, rather than the datapoint itself, means that model developers or end users of the system can intervene in terms of the concept predictions. For example, if a model predicts that two concepts are true about the input datapoint, and based on that predicts a certain target label then the model developer can determine if the final model prediction changes if one of those concepts were not true about the image.
Digging deeper, concept bottleneck models 
consist of two neural networks: the concept learning model (CLM) and the target classifier model (TLM). The CLM model assigns concept labels to data points, and the TLM predicts target labels from these predicted concept labels. Thus, the predicted concept labels provide a kind of rationale for the target label predictions. For example, a concept bottleneck model could produce predictions of the form ``I see that this image contains \textbf{wings}, a \textbf{beak}, and the background is the \textbf{sky}, so I think it is a \textbf{bird}.''

While the CBM is a compellingly simple way of forcing a neural network to base its predictions on human understandable concepts, there are a number of drawbacks that emerge when analysis of their behaviour is undertaken. For example, while the target label predictions are totally explainable as a function of the predicted concept labels, the concept labels themselves, as predicted by the CLM, can be based on spurious parts of the image as shown in \cite{margeloiu2021concept}. In the same paper, it is demonstrated that the manner of training of the TLM is important, as information about the underlying data distribution can leak into the predicted concept vectors. This can lead to a CBM performing well at tasks it should not perform well at, \textit{e.g.,} when concept labels are \textit{not} predictive of the target labels, sometimes the CBM can perform better than chance at the target label prediction. This leakage phenomenon is explored further in \cite{Lockhart2022leakage}, who give suggestions for how this can be mitigated. 
Finally, the fact that CBMs are highly dependent on all of the datapoints being labeled with concept annotations is an issue we feel has not been addressed satisfactorily in the literature. This will be the focus of the present work. Intuitively, a CBM will not perform well at the target classification task if the concept labels it is provided with are insufficient for performing a correct concept classification. In the default configuration of the CBM, there can be a catastrophic collapse in performance when the concept labels are inadequate for learning. We show this later on in this paper, and show how enabling the CLM to \textit{abstain} from providing an explanation can stop this performance degradation. In other words, we present a mechanism for enabling a CBM to present rationales for its predictions\textit{ whenever it is able to}. When it is not able to provide a rationale for its prediction, it is still able to provide a prediction. This is in contrast to the standard CBM architecture, which will provide a rationale for every prediction no matter what. 
\section{Models} 
\subsection{Concept bottleneck models}
\begin{figure}
    \centering
    \includegraphics[width=0.75\textwidth]{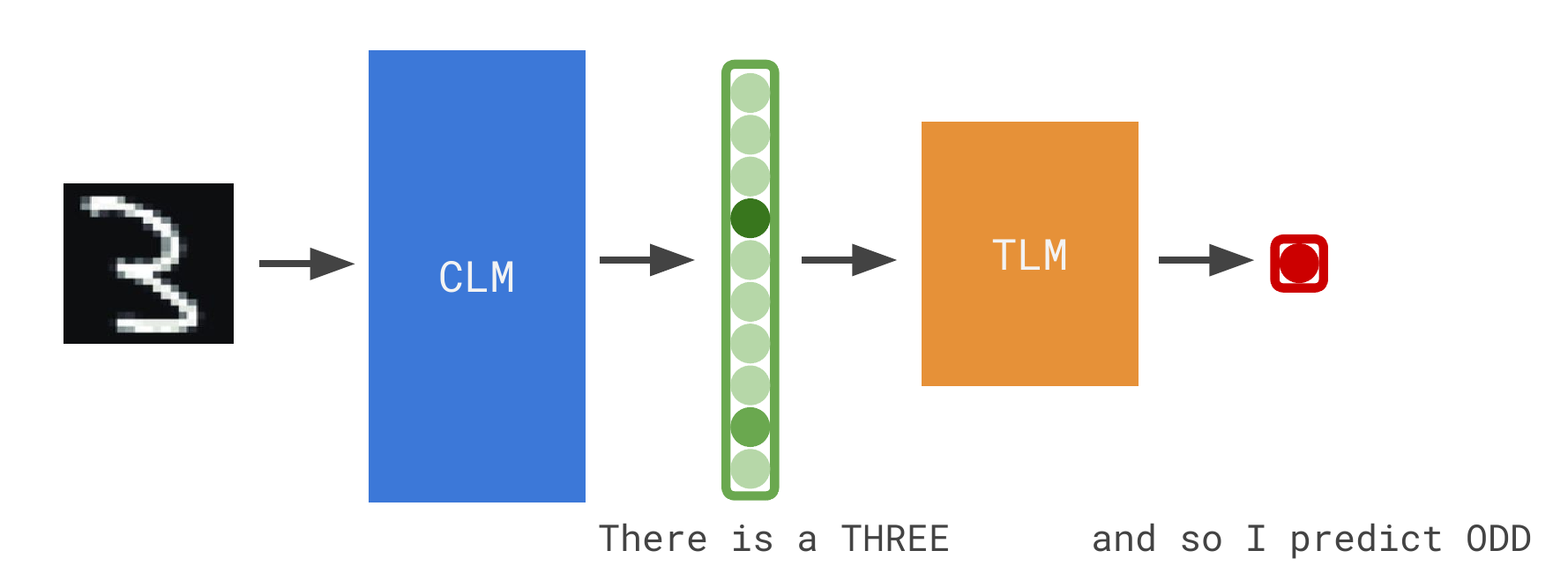}
    \caption{Concept Bottleneck Model    architecture diagram: the CLM predicts that the image is likely to depict a `3'. Based on this concept prediction, the TLM predicts that the image depicts an odd number.}
    \label{fig:cbm_arch}
\end{figure}
Conventional classification problems consist of datapoints $X_1,\dots,X_N\in \mathbb{R}^D$ sampled from some fixed data distribution $\mathcal{X}$, and target labels $y_1,\dots,y_N$ from a fixed target label set $Y=\{0,\dots,M\}$.
A classifier algorithm is trained to take a datapoint $X_i$ and produce the correct target label $y_i$.
Parametric machine learning approaches to classification problems consider parametric functions $f_\theta:\mathbb{R}^D\rightarrow Y$, (\textit{e.g.}, neural networks, generalised linear models, decision trees), optimising over parameters $\theta$ in some corresponding parameter space $\Theta$ (\textit{e.g.}, the weight space of a neural network architecture) to minimise the expected value of a loss function $\mathbb{E}[L(\theta)]=\sum_{i=0}^{N}L(\hat{y}_i,y_i)/N$, where $\hat{y}_i$ denotes the predicted target label for the $i^\text{{th}}$ datapoint $\hat{y}_i=f_\theta(X_i)$.

Concept bottleneck models operate on classification tasks where each datapoint $X_i$ has, in addition to its target label $y_i$, a vector of \textit{concept labels} $c_i=c_i^{1},\dots, c_i^{C}$. Each concept label $c_i^{j}$ is a binary variable indicating whether the $j^\text{th}$ concept is true about the datapoint in question. The CUB Birds dataset 
of \cite{wah2011caltech} is an example of such a dataset: each image of a bird comes with a target label $y$ of its species, along with 312 concept labels that encode properties of the bird itself (\textit{e.g.} \texttt{HasBillShapePointed}, \texttt{HasWingColorBlack}). 
The key idea of a concept bottleneck model (CBM) is to compose two neural network models, the \textit{concept labeling model} (CLM) and the \textit{target classifier model} (TLM). Explicitly, the CLM $f:\mathbb{R}^D\rightarrow \mathbb{R}^C$ is trained to predict concept labels from the input datapoint $X_i$. The TLM $g:\mathbb{R}^C\rightarrow \mathbb{R}^M$ is trained to predict the target labels from the predicted concept labels. Note that to cut back on notation we have neglected to include the subscripted parameters of the functions $f$ and $g$. At prediction time, the target label prediction is therefore attained by composition of these two trained models, $\hat{y}_i=g(f(X_i))$, and the concept label predictions for that datapoint $g(X_i)$ can be interpreted as rationale for this prediction. We depict the architecture of a concept bottleneck model in Figure \ref{fig:cbm_arch}.

There are several ways that a CBM can be trained. 
Following the description of \cite{CBMs}, let $L_C, L_Y:\mathbb{R}\times\mathbb{R}\rightarrow \mathbb{R}_+$ denote generic loss functions for the CLM and TLM models respectively. 
In the \textit{independent} training paradigm, the models are trained separately. That is, the CLM is trained to minimise the discrepancy between its predicted concept labels and the ground truth, $\sum_{i=0}^{N}\sum_{j=0}^C L_C([f(X_i)]_j), c_i^j)$, where we use square brackets and a subscript $j$ to denote the $j^{\text{th}}$ element of the predicted concept vector. Similarly, the TLM is trained to minimise $\sum_{i=0}^{N} L_Y(g(c_i), y_i)$. In the \textit{sequential} paradigm, the TLM is trained on the concept labels predicted by the CLM. The CLM is trained as in the independent paradigm, but the TLM is trained to minimise $\sum_{i=0}^{N} L_Y(g(f(X_i)), y_i)$. Finally, in the \textit{joint} training paradigm the losses of both CLM and TLM are minimised as a weighted sum
$L_\text{joint}=\sum_{i=0}^{N}\left( L_Y\left(g\left(f\left(X_i\right)\right), y_i\right) + \lambda\sum_{j=0}^C L_C\left(\left[f\left(X_i\right)\right]_j, c_i^j\right)\right).$

For all training paradigms, predictions for a datapoint $X_i$ are attained by first using the CLM to predict the concepts $\hat{c}_i=f(X_i)$, then passing these through the TLM to attain the target prediction $\hat{y}_i=g(\hat{c}_i)$. For the sequential and joint training paradigms a subtle issue emerges with respect to the predicted concept labels: naturally, information about the data distribution $\mathcal{X}$ can leak into these predicted concept vectors. This  seems innocuous, but \cite{mahinpei2021promises} show that this \textit{information leakage} can lead to absurd scenarios that mean the concept predictions cannot be trusted as explanations. Consider in particular their demonstration that a CBM trained to determine whether an MNIST digit is odd or even using \textit{only} the concepts \texttt{IsAThree} and \texttt{IsAFour} can perform substantially better than chance even when there are no 3 or 4 digits in the training set.

\cite{Lockhart2022leakage} dig deeper onto the mechanism by which leakage occurs, giving evidence that it can be mitigated by utilising the Monte Carlo Dropout technique of \cite{MCD} as a measure of concept uncertainty rather than simply using soft labels. In this work we will train only in the independent paradigm, and we will threshold all CAVs to attain hard concept labels when training the TLM. This ensures no leakage occurs.

\subsection{Sidecar concept bottleneck models: abstaining from explaining}
\begin{figure}
    \centering
    \includegraphics[width=0.75\textwidth]{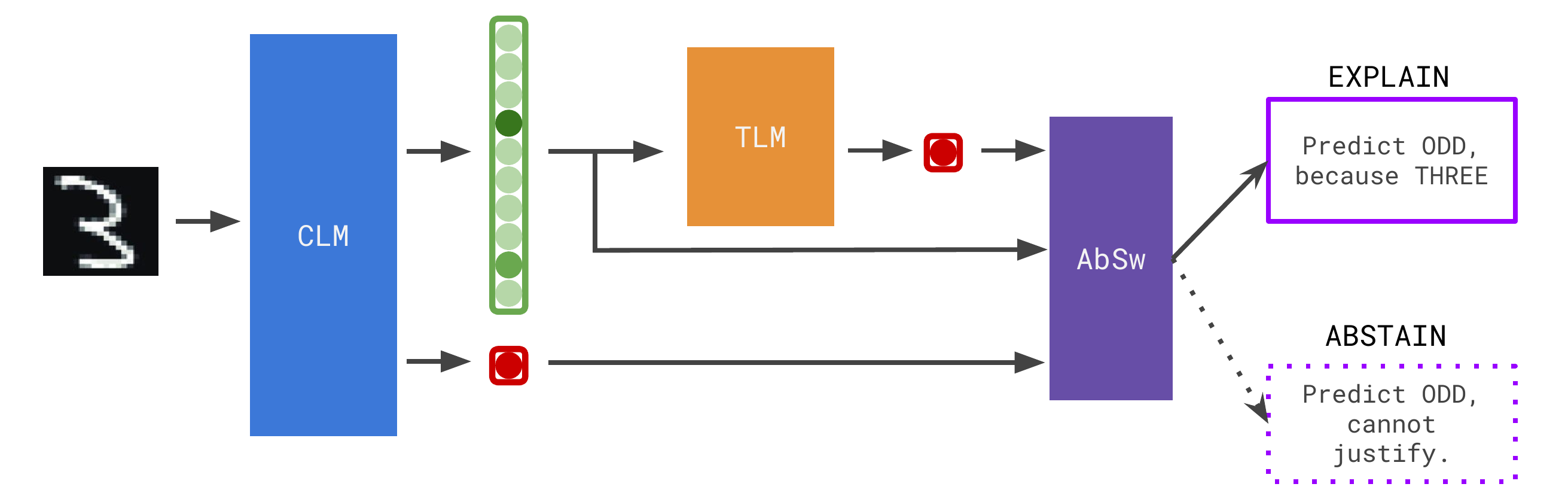}
    \caption{Architecture diagram for Sidecar Concept Bottleneck Model: the CLM predicts the concept labels, but also attempts to predict the target label. If it is uncertain of its concept predictions, then the abstention switch (AbSw) means the predicted target label of the CLM is used instead of the target label from TLM.}
    \label{fig:SCBM}
\end{figure}
Training a concept bottleneck model on a dataset with less than full concept label coverage is a bad idea: since the TLM bases its predictions solely on the predicted concepts from the CLM, if these predicted concepts are incorrect then this poor performance will cascade through to the target label predictions. 
This work focuses on how we can salvage the CBM idea on data where only a subset of datapoints are given concept labels.
Specifically, we propose augmenting concept space with the target label, training what we refer to as a \textit{Sidecar CLM}. Such a Sidecar CLM predicts the concepts and, along to the \textit{side}, the target label. The predictive uncertainty of the trained Sidecar CLM is quantified for its concept predictions.
In the case where the CLM is \textit{certain} about its concept predictions according to some criterion, the corresponding CAVs are passed through to the TLM as usual.
When the CLM is \textit{uncertain} about its concept predictions, the TLM is bypassed completely, and the target concept label as provided by the Sidecar CLM is used instead. 
In this way, the CBM is given a means of abstaining from predicting an explanation.

Formally, consider a concept augmented classification problem with datapoints $X_1,\dots,X_N\in\mathbb{R}^D$, target labels $y_1,\dots,y_N\in\{0,\dots,M\}$, and concept label vectors $c_1\dots,c_N\in\{0,1\}^{\times C}$. As elucidated in the previous section, a standard concept bottleneck model consists of a CLM trained to predict concept activation vectors for each datapoint $f:\mathbb{R}^D\rightarrow \mathbb{R}^C$, and a TLM trained to predict target labels from those predicted concept activation vectors $g:\mathbb{R}^C\rightarrow \mathbb{R}^M$. We define the \textit{Sidecar} CBM by augmenting the co-domain of the CLM $f$ to contain the target label space in addition to the concept vector space, obtaining $f_\text{S}:\mathbb{R}^D\rightarrow \mathbb{R}^C\oplus\mathbb{R}^M$. In this way, the CLM of the Sidecar CBM can predict both a concept vector and a target prediction. In a Sidecar CBM the TLM is unchanged from the TLM in the standard CBM, it maps predicted concept vectors to target label space. The key component of our architecture is the \textit{abstention switch}, which causes the bottleneck to be bypassed in the cases when the CLM is uncertain about the concept labels. The abstention switch takes the concept vector predicted by the CLM, and produces $1$ if the concept vector satisfies a criterion on how certain the prediction is, and produces $0$ otherwise. If the vector fails this test then the CLM is deemed to be too uncertain about its prediction to use its CAV as an explanation, and the target prediction of the CLM is used instead. The uncertain CAV is not passed to the TLM for a target prediction. Explicitly, the abstention switch is a function $a:\mathbb{R}^C\rightarrow\{0,1\}$ that decides whether to use the predicted CAV from CLM. Abusing notation, we refer to the target prediction portion of the output of the Sidecar CLM $f_S$ as the quantity $[f_S(\cdot)]_t$, and the concept prediction portion of the output as $[f_S(\cdot)]_c$.
\newpage
Thus, given a trained Sidecar CLM $f_S$, a trained TLM $g$, and an abstention switch $a$ we attain our target prediction as 
\begin{align*}\hat{y}_i = g([f_S(X_i)]_c)^{a([f_S(X_i)]_c} + [f_S(X_i)]_t^{1-a([f_S(X_i)]_c)}.\end{align*}
Likewise, the concept prediction, or \textit{explanation} for the target prediction is $[f_S(X_i)]_c$ if $a([f_S(X_i)]_c=1$, and NULL otherwise (providing NULL meaning the model has abstained from providing rationale).

The architecture of the Sidecar CBM is illustrated in Figure \ref{fig:SCBM}. 
The abstention switch $a$ can be any measure of predictive uncertainty. For our experiments we need a measure of uncertainty suitable for multi-label classification problems. Note that while uncertainty can be estimated in the single label setting by a wide variety of techniques, \textit{e.g.}, the Shannon entropy of the softmax distribution, this idea becomes non-trivial when multiple labels are involved. Indeed, one could use the joint Shannon entropy, but the complexity of calculating this quantity grows exponentially in the number of labels. For our purposes, we found that it is suitable to treat the predicted CAV as a probability vector, passing each logit of the final layer through a sigmoid activation function. Then, the predictive uncertainty of the multi-label prediction is simply how far away from unit probability the predicted most likely class is. The abstention switch can then be defined as a threshold on that uncertainty quantification. That is, we define $a_\epsilon(v)$ to be equal to $1$ if $\max(v) > \epsilon$, and $0$ otherwise.
We note that this abstention switch idea is defined to be very general, and can be replaced by another means of testing for uncertainty as needed.
In the next section we test the performance of using abstention.

\section{Experiments} 
In order to determine how well our models perform, we need to conduct experiments on datasets with concept labels and target labels. To this end, we construct our own datasets from existing benchmark datasets. Specifically, we consider data with target labels $Y$, then define our own concept labels based on those target labels. Each modified \textit{concept} dataset will have another target label $Y'$. For example, we consider the problem of determining if an \texttt{MNIST} digit is odd or even: the target $Y'$ of this concept based dataset is a function of the original target label $Y$, and the concept labels are the original labels. The CBM will then learn to produce predictions of whether the digit is odd or even, based on, and justifiable by, the prediction of the digit's identity. Let us now explain the datasets we use in more depth.
\subsection{Datasets}
\texttt{ParityMNIST} as used by \cite{mahinpei2021promises} is based on the \texttt{MNIST} dataset of handwritten digits (\cite{lecun1998gradient}). Let $X_i$ be a datapoint from the latter and let $y_i$ be its label. The corresponding \texttt{ParityMNIST} label for that datapoint is $0$ if $y_i$ corresponds to an even digit, and $1$ if $y_i$ is odd -- the task is to determine if the \texttt{MNIST} digit is an odd or even number. Concept labels for this task are the original \texttt{MNIST} target labels, that is, $c_i^j = 1\text{ if } y_i = j$, and  $0 \text{ otherwise}$.
\texttt{AliveCIFAR10} is based on \texttt{CIFAR-10} (\cite{krizhevsky2009learning}). \texttt{CIFAR-10} is comprised of images of airplanes, cars, ships, trucks, cats, birds, deer, dogs, frogs, and horses. The first four of these labels are entities that are not alive, the latter six are living creatures. Thus, the \texttt{AliveCIFAR10} dataset is a binary classification task to distinguish the former and the latter. Like the previous dataset, the concept labels are the original classification task labels. 
\texttt{InOutFashionMNIST} is based on the \texttt{FashionMNIST} dataset of \cite{xiao2017fashion}, which consists of pictures of clothing . The inspiration for this dataset comes from a question that faces us all: which garment in our wardrobe is appropriate for outdoor tasks, and which are more general purpose? Thus, we derive the \texttt{InOutFashionMNIST} task by splitting the garment image classes into those that the authors would add to their outfit before leaving the house to come to the office (\textit{e.g.}, coat, shoes, bag), and those which we deem to be general purpose garments worn as part of a normal outfit inside and outside (\textit{e.g.}, shirt, dress, trousers).
In Table \ref{tab:datasets} we formalise the concept and target labels for these problems.
\begin{table}[]
    \centering
    \tiny
    \setlength{\tabcolsep}{3pt}
    \begin{tabular}{|c|c|c|c|c|c|c|c|c|c|c|}
    \hline
    \textbf{Dataset} & \multicolumn{10}{c|}{\textbf{Target labels}}\\
    \hline
    \texttt{MNIST} & 0&1&2&3&4&5&6&7&8&9\\
    \texttt{ParityMNIST} & 0&1&0&1&0&1&0&1&0&1\\
    \hline
    \texttt{FashionMNIST} & T-shirt/top & Trouser & Pullover & Dress & Coat & Sandal & Shirt & Sneaker & Bag & Ankle boot \\
    \texttt{InOutFashionMNIST} & 0 & 0 & 0 & 0 & 1 & 1 & 0 & 1 & 1 & 1\\
    \hline
    \texttt{CIFAR10} & airplane & automobile & bird & cat & deer & dog & frog & horse & ship & truck \\
    \texttt{AliveCIFAR10} & 0 & 0 & 1 & 1 & 1 & 1 & 1 & 1 & 0 & 0\\
    \hline
    \end{tabular}
    \caption{Tabular description of each of the concept augmented datasets we use for the experiments.}
    \label{tab:datasets}
\end{table}

We are interested in three quantities of a concept bottleneck model: the \textit{target accuracy}, the \textit{explanation accuracy}, and the \textit{explanation coverage}. The first is to ascertain if the target label predictions are correct. Since the datasets we consider are relatively balanced, we use raw accuracy. The second is to determine the quality of the concept predictions. We are interested in the possibility of using these concept predictions as justifications or \textit{explanations} for the model prediction. Hence, it is important that the explanations are correct. Finally, since the Sidecar CBM is capable of abstaining from giving an explanation, it is crucial that we understand how often it is doing so. By the design of the Sidecar CBM architecture, target accuracy should stay high. However, if this comes at the expense of the number of explained predictions produced then we would like to know.

For all experiments, the CLM models were fit by finetuning ResNet-18 (\cite{he2016deep}) with the Adam optimiser (\cite{kingma2014adam}), learning rate of $0.0001$, and batch size of $100$. We ran the training until we deemed the model had converged using a very simple early stopping criterion: if the binary cross entropy loss on our validation set didn't get better for more than three epochs then we stopped training. The TLM models were fully connected models, with a single hidden layer of 128 neurons. Since we trained the CLM and TLM independently, the TLM models were trained to predict the target labels from the ground truth concept labels and converged very quickly to a very low validation loss. In order to judge the predictive uncertainty of the Sidecar CLM, we used the sigmoid activations as probabilities to feed in to the threshold abstention switch we describe above. The threshold we used for abstention was $0.75$. That is, if the highest predicted probability for a concept label was less than $0.75$, then the CLM abstains from predicting its CAV.

We are interested in the performance of the models on concept datasets with missing concept labels. Therefore, we randomly removed concept labels from the training datasets according to various fixed probabilities to test model performance. 
\subsection{Results}
\begin{figure}
    \centering
    \includegraphics[width=0.75\textwidth]{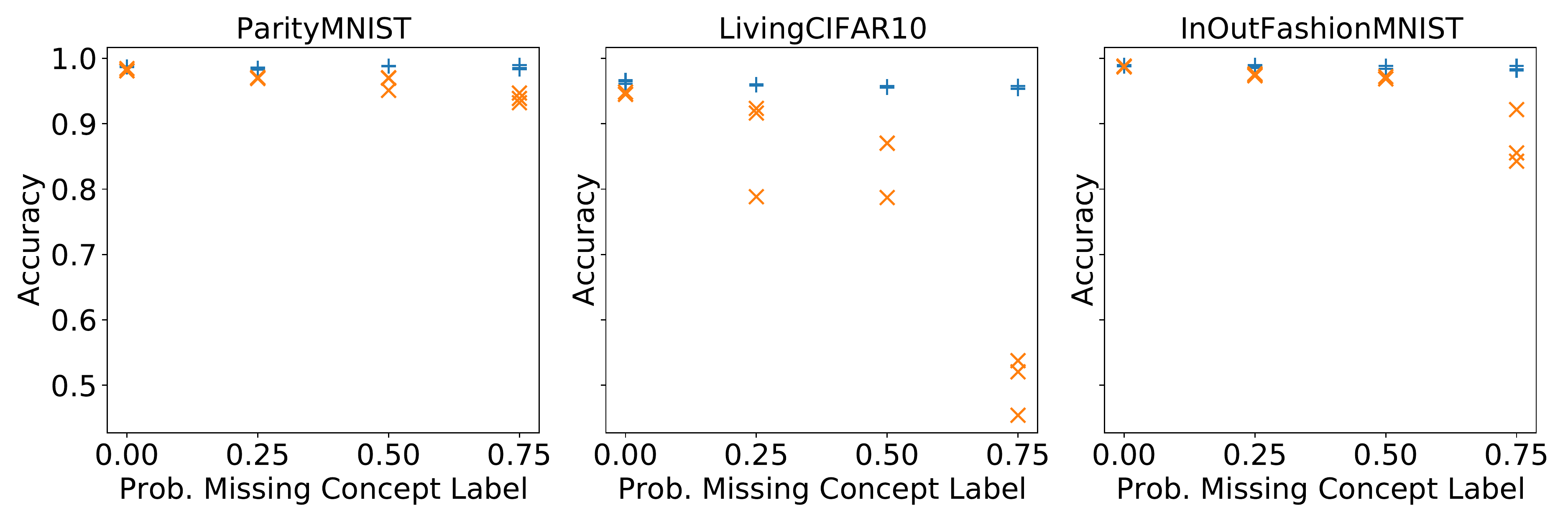}
    \caption{Target label accuracy for standard Concept Bottleneck Model (orange crosses), compared to target label accuracy for Sidecar Concept Bottleneck Model (blue pluses).}
    \label{fig:acc_plot}
\end{figure}

In Table \ref{tab:results1} we show average test set \textit{target accuracy} for the CBM, and our Sidecar CBM, on the three datasets we outlined previously \texttt{LivingCIFAR10}, \texttt{ParityMNIST}, and \texttt{InOutFashionMNIST}. We vary the probability of missing concept labels from $0.0$ to $0.75$. It is informative to compare these accuracy numbers with the accuracy of a model trained to directly predict the target label for each dataset. To this end, for each dataset we finetuned a ResNet-18 model (same as the CLM for each CBM) to directly predict the target label, with the same hyperparameters as specified for CLM training. Repeating this three times, we obtained $0.986\pm 0.003$ for \texttt{InOutFashionMNIST}, $0.954\pm 0.002$ for \texttt{LivingCIFAR10}, and 
$0.985\pm 0.002$ for \texttt{ParityMNIST}. These results are illustrated in the plots in Figure \ref{fig:acc_plot}.

In Table \ref{tab:results2} we show \textit{explanation accuracy}, that is, the F1 score of the task of predicting the correct concept labels. Finally, \ref{tab:results3} shows the \textit{explanation coverage}, literally the proportion of the test set datapoints that the model provided an explanation for (did not abstain). Since the standard CBM cannot abstain, the number is always $1$ on those rows.
All experiments were run three times to obtain the averaged numbers shown in the tables.

We see that there is a substantial performance degradation for the standard CBM as the proportion of data points with concept labels decreases. Compare this with the only very slight performance degradation of our Sidecar CBM. In Table \ref{tab:results2} we show the explanation accuracy, where again we see a dramatic decrease in explanation accuracy for the standard concept bottleneck model. We see the performance actually increases for the sidecar model, however from looking at Table \ref{tab:results3} we see that the explanation coverage drops off as missing concept labels increases. 
\section{Discussion} 
We have presented a neural network architecture that provides predictions that are justified in terms of human provided concepts. Based on the \textit{concept bottleneck} approach to classification of \cite{CBMs}, there are two components to the architecture: the concept labeling model (CLM) which is trained to determine which of a fixed number of \textit{concepts} are true about the input datapoint; while the target labeling model (TLM) is trained to produce target labels from the predicted concept labels. In this way, the target predictions are based solely on the predicted concepts, and so the target predictions can be presented with these concept labels as a rationale or \textit{explanation} for why that target label was produced. Furthermore, since the target predictions are based solely on the concept predictions and not the datapoint itself, users of the model can intervene on concepts to determine how that would change the target prediction.
\begin{table}[h!]
\tiny
\centering
\begin{tabular}{llllll}
\toprule
            & Prob. Missing Concept &              0.00 &              0.25 &              0.50 &              0.75 \\
Dataset Name & Sidecar CLM &                   &                   &                   &                   \\
\midrule
InOutFashionMNIST & False &  $0.988\pm 0.001$ &  $0.975\pm 0.001$ &   $0.97\pm 0.002$ &  $0.873\pm 0.035$ \\
            & True  &  $0.989\pm 0.001$ &  $0.987\pm 0.002$ &  $0.987\pm 0.002$ &  $0.984\pm 0.003$ \\
LivingCIFAR10 & False &  $0.946\pm 0.002$ &  $0.876\pm 0.062$ &  $0.843\pm 0.039$ &  $0.504\pm 0.036$ \\
            & True  &  $0.964\pm 0.002$ &  $0.959\pm 0.001$ &  $0.956\pm 0.001$ &  $0.955\pm 0.002$ \\
ParityMNIST & False &  $0.982\pm 0.002$ &   $0.97\pm 0.001$ &  $0.964\pm 0.009$ &  $0.939\pm 0.006$ \\
            & True  &    $0.987\pm 0.0$ &  $0.985\pm 0.001$ &    $0.988\pm 0.0$ &  $0.986\pm 0.003$ \\
\bottomrule
\end{tabular}
\caption{Average, and standard deviation of accuracy for the target task.}
\label{tab:results1}
\end{table}
\begin{table}[h!]
\tiny
\centering
\begin{tabular}{llllll}
\toprule
            & Prob. Missing Concept &              0.00 &              0.25 &              0.50 &              0.75 \\
Dataset Name & Sidecar CLM &                   &                   &                   &                   \\
\midrule
InOutFashionMNIST & False &     $0.97\pm 0.0$ &  $0.946\pm 0.006$ &   $0.94\pm 0.002$ &  $0.763\pm 0.016$ \\
            & True  &  $0.987\pm 0.001$ &    $0.993\pm 0.0$ &  $0.997\pm 0.001$ &      $1.0\pm 0.0$ \\
LivingCIFAR10 & False &  $0.733\pm 0.002$ &  $0.624\pm 0.008$ &  $0.506\pm 0.021$ &  $0.206\pm 0.066$ \\
            & True  &  $0.908\pm 0.003$ &  $0.936\pm 0.003$ &   $0.97\pm 0.003$ &  $0.948\pm 0.017$ \\
ParityMNIST & False &  $0.967\pm 0.003$ &  $0.943\pm 0.003$ &  $0.925\pm 0.014$ &  $0.878\pm 0.014$ \\
            & True  &  $0.988\pm 0.001$ &  $0.993\pm 0.001$ &  $0.997\pm 0.001$ &      $1.0\pm 0.0$ \\
\bottomrule
\end{tabular}

\caption{Average, and standard deviation of F1 score for explanation accuracy.}
\label{tab:results2}
\end{table}
\begin{table}[h!]
\tiny
\centering
\begin{tabular}{llllll}
\toprule
            & Prob. Missing Concept &             0.00 &             0.25 &             0.50 &             0.75 \\
Dataset Name & Sidecar CLM &                  &                  &                  &                  \\
\midrule
InOutFashionMNIST & False &      $1.0\pm 0.0$ &      $1.0\pm 0.0$ &      $1.0\pm 0.0$ &      $1.0\pm 0.0$ \\
            & True  &  $0.966\pm 0.003$ &  $0.905\pm 0.017$ &  $0.453\pm 0.064$ &  $0.002\pm 0.002$ \\
LivingCIFAR10 & False &      $1.0\pm 0.0$ &      $1.0\pm 0.0$ &      $1.0\pm 0.0$ &      $1.0\pm 0.0$ \\
            & True  &  $0.653\pm 0.059$ &  $0.423\pm 0.013$ &   $0.11\pm 0.025$ &  $0.005\pm 0.005$ \\
ParityMNIST & False &      $1.0\pm 0.0$ &      $1.0\pm 0.0$ &      $1.0\pm 0.0$ &      $1.0\pm 0.0$ \\
            & True  &  $0.963\pm 0.005$ &   $0.88\pm 0.012$ &  $0.419\pm 0.071$ &  $0.001\pm 0.001$ \\
\bottomrule
\end{tabular}
\caption{Proportion of datapoints in test set that the model provided an explanation for. Note that the standard CBM always provides an explanation, so proportion is always $1$.}
\label{tab:results3}
\end{table}

Our contribution is to equip the CLM component in the above with the ability to abstain from producing a concept prediction. Crucially, we do so in a way that enables the full CBM to still produce a target prediction when the CLM abstains. A big issue with the standard CBM approach is that a poor quality CLM will produce incorrect concept predictions, and since the TLM component is trained only on \textit{correct} concept labels, these incorrect concept predictions can be out of distribution. This causes major isses, as we have seen in the experiments we conducted and discussed above. A failure of the CLM causes a cascade in bad target predictive performance through to the TLM component. Our sidecar approach to CBM is to train the CLM on concept labels \textit{and} target labels. That way, if the CLM is uncertain about the concept labels, then we can ignore them and simply take its target prediction as the CBM's target prediction. The bad concept predictions are not fed through the model, and the performance of the CBM on the target task does not degrade. Otherwise, if the CLM is certain enough about its concept predictions then these can be fed through to the TLM as usual.

Looking at the results in Table \ref{tab:results1}, our architecture outperforms the standard CBM in the setting where all concept labels are present in the training data (probability of missing concept label is $0.0$). In this experiment we see that the explanation coverage is also high (see Table \ref{tab:results3}). Additionally, we note that when the proportion of missing concept labels increases, the standard CBM performs poorly, while the Sidecar CBM is able to maintain high performance accuracy. The ability to abstain gives our architecture a robustness advantage over the standard CBM by not requiring an explanation for every prediction.
Essentially, we are letting the model explain its own predictions, but only when it is happy to do so. If it deems itself incapable of providing a good rationale for a prediction, it can just provide the prediction itself with \textit{no} rationale. The end user of the system can therefore make a judgement call as to whether or not the prediction should be used, depending on the task at hand. Another crucial aspect of the above is that we have maintained the ability of the end user to intervene on concept predictions for the datapoints for which a rationale is provided. 
\section*{Acknowledgements}
This paper was prepared for informational purposes by the Artificial Intelligence Research group of JPMorgan Chase \& Co and its affiliates (``J.P. Morgan''), and is not a product of the Research Department of J.P. Morgan.  J.P. Morgan makes no representation and warranty whatsoever and disclaims all liability, for the completeness, accuracy or reliability of the information contained herein.  This document is not intended as investment research or investment advice, or a recommendation, offer or solicitation for the purchase or sale of any security, financial instrument, financial product or service, or to be used in any way for evaluating the merits of participating in any transaction, and shall not constitute a solicitation under any jurisdiction or to any person, if such solicitation under such jurisdiction or to such person would be unlawful.   

\bibliographystyle{plainnat}
\bibliography{refs}

\end{document}